\documentclass[conference]{IEEEtran}
\IEEEoverridecommandlockouts
\usepackage{cite}
\usepackage{amsmath,amssymb,amsfonts}
\usepackage{algorithmic}
\usepackage{graphicx}
\usepackage{textcomp}
\usepackage[table]{xcolor}
\usepackage{hyperref}
\def\BibTeX{{\rm B\kern-.05em{\sc i\kern-.025em b}\kern-.08em
    T\kern-.1667em\lower.7ex\hbox{E}\kern-.125emX}}
\begin{document}

\title{Illicit object detection in X-ray images using Vision Transformers \thanks{Research was possible due to the funding from the European Union's Horizon Europe research and innovation programme under grant agreement No 101073876 (Ceasefire)}}

\author{
\IEEEauthorblockN{Jorgen Cani}
\IEEEauthorblockA{\textit{Department of Informatics and Telematics} \\
\textit{Harokopio University of Athens}\\
Athens, Greece \\
cani@hua.gr}
\and
\IEEEauthorblockN{Ioannis Mademlis}
\IEEEauthorblockA{\textit{Department of Informatics and Telematics} \\
\textit{Harokopio University of Athens}\\
Athens, Greece \\
imademlis@hua.gr}
\and
\IEEEauthorblockN{Adamantia Anna Rebolledo Chrysochoou}
\IEEEauthorblockA{\textit{Department of Informatics and Telematics} \\
\textit{Harokopio University of Athens}\\
Athens, Greece \\
adamantia.reb@hua.gr}
\and
\IEEEauthorblockN{Georgios Th. Papadopoulos}
\IEEEauthorblockA{\textit{Department of Informatics and Telematics} \\
\textit{Harokopio University of Athens}\\
Athens, Greece \\
g.th.papadopoulos@hua.gr}
}

\maketitle



\begin{abstract}
Illicit object detection is a critical task performed at various high-security locations, including airports, train stations, subways, and ports. The continuous and tedious work of examining thousands of X-ray images per hour can be mentally taxing. Thus, Deep Neural Networks (DNNs) can be used to automate the X-ray image analysis process, improve efficiency and alleviate the security officers' inspection burden. The neural architectures typically utilized in relevant literature are Convolutional Neural Networks (CNNs), with Vision Transformers (ViTs) rarely employed. In order to address this gap, this paper conducts a comprehensive evaluation of relevant ViT architectures on illicit item detection in X-ray images. This study utilizes both Transformer and hybrid backbones, such as SWIN and NextViT, and detectors, such as DINO and RT-DETR. The results demonstrate the remarkable accuracy of the DINO Transformer detector in the low-data regime, the impressive real-time performance of YOLOv8, and the effectiveness of the hybrid NextViT backbone.
\end{abstract}

\begin{IEEEkeywords}
Object Detection, X-rays, Vision Transformers, Deep Neural Networks
\end{IEEEkeywords}

\section{Introduction}
The detection of concealed illicit items using X-ray images is a common security procedure \cite{mademlisVisualInspectionIllicit2023,batsis2023illicit}. X-ray scanners are a non-invasive way to examine suspicious containers, such as mailed parcels or luggage, in places where security is of utmost importance, such as airport terminals, customs offices, post offices, etc. \cite{hassanCascadedStructureTensor2020}. The detection process involves X-ray machines directing beams of high-energy radiation towards the object being scanned. The varying densities of materials within the object cause the X-rays to be absorbed at different rates. The resulting high resolution X-ray images, where denser materials appear lighter and less dense materials appear darker, are manually inspected for the presence of illicit items (e.g., firearms) by security officers in real-time \cite{meryXRayBaggageInspection2020}.

However, X-ray scans present certain shortcomings that perpetrators can exploit for concealing contraband, such as the occlusion of layered objects, a cluttered environment, similarity of different objects, as well as certain material properties which may impact the image appearance \cite{akcayAutomaticThreatDetection2022a}. Additionally, heavy traffic during rush hours may mentally overload the security officers, causing poor decisions during manual inspection. Hence, efficient and well-performing automated solutions are necessary for overcoming such issues.

Recent advances in object detection and in the exploitation of multiple modalities \cite{rodis2023multimodal} have led to promising X-ray image analysis algorithms \cite{rafieiComputerVisionXRay2023}. Deep Neural Network (DNN) architectures such as Convolutional Neural Networks (CNNs) \cite{krizhevskyImageNetClassificationDeep2012} and Vision Transformers \cite{dosovitskiyImageWorth16x162020} are the state-of-the-art approaches in general object detection, where the DNN analyzes its input image and outputs the bounding boxes/Regions-of-Interests (RoIs) of all objects visible on it, in pixel coordinates, along with the corresponding class labels. However, when it comes to detecting illicit items in X-ray scan images, the more recent and powerful Transformer architectures, which rely on the attention mechanism \cite{vaswaniAttentionAllYou2017}, have not been extensively employed. Relevant literature is still mostly dominated by CNN-based approaches \cite{ganYOLOCIDImprovedYOLOv72023,maOccludedProhibitedObject2023,renLightRayLightweightNetwork2022}, similarly to visual firearms recognition in the RGB domain \cite{konstantakos2023selfSupervised,mademlis2023invisible}.

In order to address these gaps in the literature, this paper attempts to systematically assess the capabilities of Vision Transformers at the task of illicit item detection in X-ray images. A hybrid architecture that internally combines convolutions and the attention mechanism is also being evaluated. Thus, extensive experimental comparisons are conducted using the SWIN \cite{liuSwinTransformerHierarchical2021} and Next-ViT \cite{liNextViTNextGeneration2022} deep neural backbones, as well as the DINO \cite{zhangDINODETRImproved2022} and RT-DETR \cite{lvDETRsBeatYOLOs2023} detection heads. Additionally, YOLOv8 \cite{jocherUltralyticsYOLOv82023} is included as a baseline CNN-based one-stage object detector. Evaluation is conducted by pretraining on SIXRay \cite{miaoSIXrayLargescaleSecurity2019}, i.e., a relevant, large-scale public dataset, and then finetuning the object detectors on CFray, i.e., a smaller, custom dataset containing X-ray images of parcels containing firearms or related objects.

The remainder of this paper is organized in the following manner. Section \ref{related-work} briefly surveys the state-of-the-art regarding illicit item detection in X-ray images using DNNs. Section \ref{obj-det-methods} presents the specific deep neural architectures that are being evaluated in this paper. Subsequently, Section \ref{evaluation} discusses the evaluation datasets, process and results. Finally, Section \ref{conclusion} draws conclusions from the conducted study.


\section{Related Work}
\label{related-work}
A wide range of the methods have been employed over the years for detecting illicit items in X-ray scan images. The method in \cite{akcayTransferLearningUsing2016} adopts a complex CNN pretrained in natural RGB images and finetunes it at the X-ray domain, in an attempt to address the common issue of DNN overfitting to limited data. This is a significant problem in relevant literature, since it is difficult to construct large-scale X-ray image datasets due to the high cost and low availability of X-ray scanners. In addition, the percentage of ``positive" objects (meaning parcels or luggage that indeed contain illicit items) is typically very low in such datasets, rendering effective DNN training even more difficult.

Common DNNs for general object detection have repeatedly been evaluated in this domain, including Faster R-CNN \cite{renFasterRCNNRealTime2015}, Mask R-CNN \cite{heMaskRCNN2017}, and RetinaNet \cite{linFocalLossDense2017}. However, the standard approach is to use real-time one-stage object detectors such as YOLO \cite{redmonYouOnlyLook2015} or SSD \cite{liuSSDSingleShot2016}, due to their significantly higher inference speed. LightRay \cite{renLightRayLightweightNetwork2022}, an alternative lightweight object detector, is a modification of YOLOv4 designed to detect small illicit items detection in complex environments. It consists of a MobileNetV3 \cite{howardSearchingMobileNetV32019} backbone CNN, a feature enhancement network and a Lightweight Feature Pyramid Network (LFPN) \cite{linFeaturePyramidNetworks2017}, that obtains information about objects at different scales, and it also includes a Convolutional Block Attention Module (CBAM) \cite{wooCBAMConvolutionalBlock2018} for refining feature maps using a spatial attention mechanism.

A different approach is taken in \cite{shaoExploitingForegroundBackground2022a}, where a new mechanism called Foreground and Background Separation (FBS) is proposed to separate illicit items from complex or cluttered backgrounds. This is accomplished by using a feature extraction DNN combined with Spatial Pyramid Pooling (SPP) and a Path Aggregation Network to extract high-level features. These feature maps serve as input to two neural decoders that simultaneously reconstruct the background and foreground. An attention module then directs the model's focus to the foreground objects.

YOLO-CID \cite{ganYOLOCIDImprovedYOLOv72023} is a modified YOLOv5 architecture that focuses on real-time performance. The architecture is enhanced with the Stem \cite{wangPeleeRealTimeObject2019} and CGhost \cite{hanGhostNetMoreFeatures2020} neural modules. Thus, it achieves accuracy competitive with respect to the baseline, but with a smaller model complexity and higher inference speed.

\section{Object Detection Methods}
\label{obj-det-methods}

This section details the selected deep neural architectures, starting from the backbone networks and then moving on to the object detection heads.

\subsection{Backbone Networks}

SWIN \cite{liuSwinTransformerHierarchical2021} is a commonly used hierarchical Vision Transformer that attempts to enhance scale-invariance in image analysis. It is composed of the Swin Transformer block, which replaces the standard multi-head self attention (MSA) block. SWIN is a strong backbone network option for object-detection methods, as seen in \cite{zhangDINODETRImproved2022} and \cite{zongDETRsCollaborativeHybrid2023}. Four variants of the SWIN Transformer network exist: tiny, small, base and large. This paper utilizes the base configuration of SWIN, consisting of 88 million parameters.

\begin{figure*}[h]
    \centering
    \includegraphics[width=1.22\columnwidth]{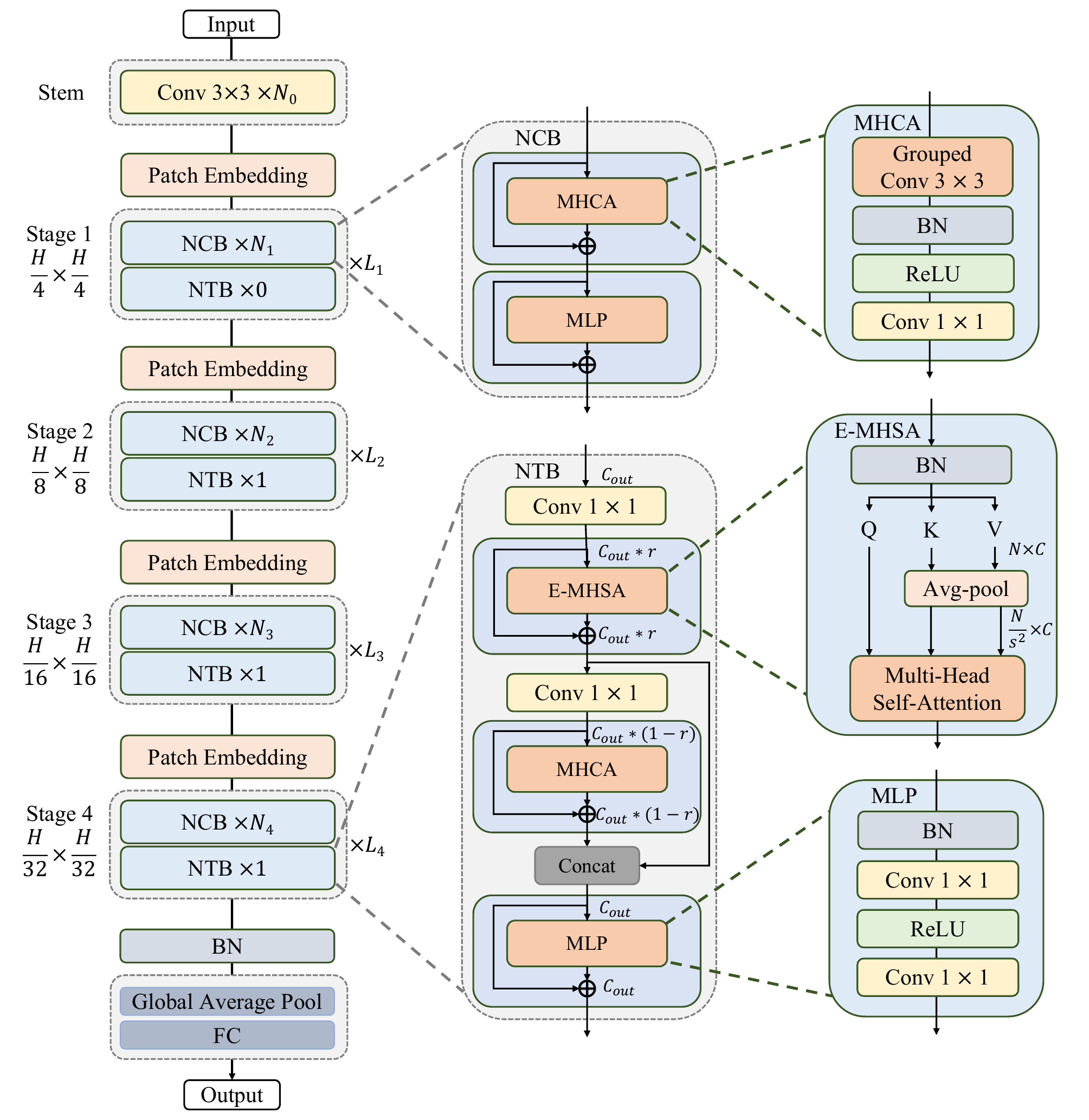}
    \caption{The architecture of the NextViT-s backbone. Image from \cite{liNextViTNextGeneration2022}.}
    \label{fig:nextvit}
\end{figure*}

NextViT \cite{liNextViTNextGeneration2022} is a backbone network that aims for a latency/accuracy trade-off. ViTs have great results, but are slow to infer. In contrast, CNNs are fast to infer but inferior to ViTs in terms of performance. NextViT introduces the Next Convolution Block (NCB) and Next Transformer Block (NTB), which, respectively, capture both local and global information. The architecture is primarily designed for industrial/embedded use-cases in mind, such as security inspection devices in airport terminals, subways, etc. It relies on the Next Hybrid Strategy (NHS), which creates four stages of various NCB and NTB blocks, where each stage's number of blocks depends on the configuration of the model. The available configurations are small, base and large.

This paper utilizes the small variant of NextViT, which is depicted in Fig. \ref{fig:nextvit}. The figure shows that NextViT-s consists of four stages, with the final layers being the closest to the final output. Overall, there are 24 layers:
\begin{enumerate}
    \item Layers 0-4 are stem layers.
    \item Layers 4-6 are NCB components, named Stage 1.
    \item Layers 7-10 contain three NCB components, followed by an NTB component, named Stage 2.
    \item Layers 11-20 contain four NCB component, followed by an NTB component, which in turn, is followed by four other NCB components and one final NTB component, named Stage 3.
    \item Layers 21-23 contain two NCB components and one NTB component, named Stage 4, which is the final stage.
\end{enumerate}

\subsection{Detection Head Networks}
\begin{figure*}
    \centering
    \includegraphics[width=1.5\columnwidth]{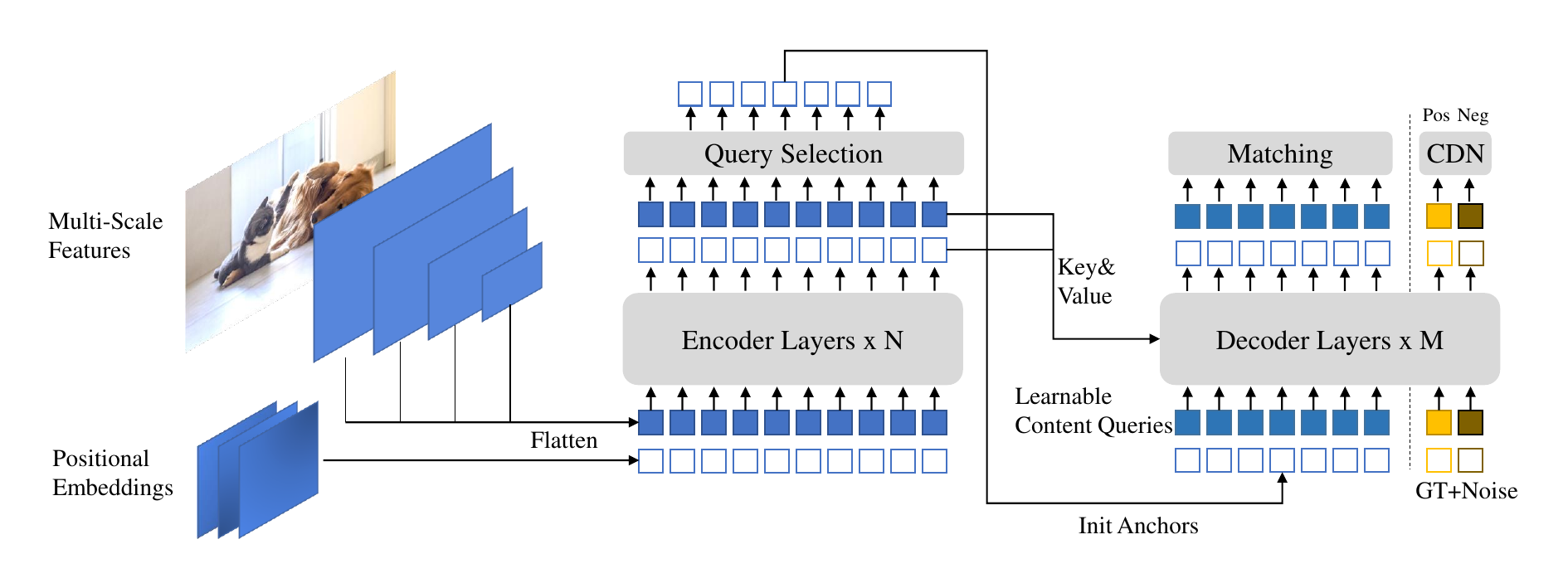}
    \caption{The DINO architecture. Image from \cite{zhangDINODETRImproved2022}.}
    \label{fig:dino-arch}
\end{figure*}

Detection Transformer (DETR) \cite{carionEndtoEndObjectDetection2020} is a Transformer-based detector, which is placed after a CNN backbone. It handles object detection as a set prediction task. DETR is not only anchor-free, but also eliminates the need of handcrafted components (e.g., the Non-Maximum Suppression algorithm). \textbf{D}ETR with \textbf{I}mproved de\textbf{N}oising anch\textbf{O}r boxes (DINO) \cite{zhangDINODETRImproved2022} is a DETR-like method that boasts impressive performance gains by accumulating various minor enhancements over the base DETR and by reintroducing the usage of anchor boxes in a Transformer setting. The goal of DINO is to avoid duplicate bounding box outputs that correspond to a single ground-truth object, which is achieved by adding two different types of noise to a single ground-truth RoI during training. Fig. \ref{fig:dino-arch} depicts the DINO architecture.

\textbf{R}eal-\textbf{T}ime \textbf{DE}tection \textbf{TR}ansformer (RT-DETR) \cite{lvDETRsBeatYOLOs2023} is the first real-time end-to-end DETR-like object detector. The architecture consists of a backbone, the Efficient Hybrid Encoder, an IoU-aware Query selector, and a Decoder \& Head final component. The object queries in DETR-like models are a set of learnable embeddings, which are optimized by the decoder component and are mapped to the classification values and bounding boxes by the prediction head. These queries however are hard to interpret and optimize. RT-DETR introduces the IoU score into the objective loss function of the classification branch, which leads to higher quality encoder features. In common implementations, RT-DETR is typically combined with the HGNetV2 CNN backbone network, with residual connections at various stages linking the backbone with the detection head.

\textbf{Y}ou \textbf{O}nly \textbf{L}ook \textbf{O}nce (YOLO) \cite{redmonYouOnlyLook2015} is a fast, anchor-based, single-stage object detector that performs object localization and classification using a single CNN. YOLO is an efficient algorithm that has gained popularity due to its speed and accuracy. Its architecture consists of a backbone network, a neck network, and a final prediction head. The latest iteration of the architecture, YOLOv8 \cite{jocherUltralyticsYOLOv82023}, is used in this paper.

Many variants of the model exist (nano, small, medium, large, extra-large; n, s, m, l, x); this paper exploits YOLOv8-l. Vanilla YOLOv8 implementations cooperate with a variant of the YOLO-specific DarkNet CNN backbone. Residual connections at various stages are exploited to connect this default CNN backbone to the detection neck and head. There are a few exceptions in the literature to the privileged use of DarkNet with YOLO. For instance, AG-YOLO \cite{linAGYOLORapidCitrus2024} leverages NextViT for optimal performance in a citrus fruit detection task, thus better addressing occlusion problems.

\section{Experimental Evaluation}
\label{evaluation}
This section details the experimental evaluation datasets and metrics, the exact experimental setup and algorithmic configuration, as well as the obtained results.

\subsection{Experimental Datasets}
\label{datasets}
This paper has utilized the public SIXray \cite{miaoSIXrayLargescaleSecurity2019} dataset for pretraining the competing DNNs, before subsequently finetuning them on the smaller, custom CFray dataset.

The SIXray \cite{miaoSIXrayLargescaleSecurity2019} dataset is a publicly available dataset that serves as an object detection benchmark for identifying illicit goods on X-ray scan images of travel luggage. It consists of $1,059,231$ X-ray images taken at subway stations, where $8,929$ illicit items are annotated. There are $6$ classes contained in the dataset: ``gun", ``knife", ``wrench", ``pliers" and ``scissors". The final class indicates the absence of illicit objects, named ``negative". The authors split the dataset into three subsets, named SIXray10, SIXray100 and SIXray1000, where the number indicates the ratio of negative to positive images. SIXRay-D \cite{nguyenMoreEfficientSecurity2022} is a modified version of the dataset, utilizing a cropping scheme that preserves only the task-related regions of each X-ray image. This paper exploits the SIXRay-D version of SIXray10.

\begin{figure}
    \centering
    \includegraphics[width=1\linewidth]{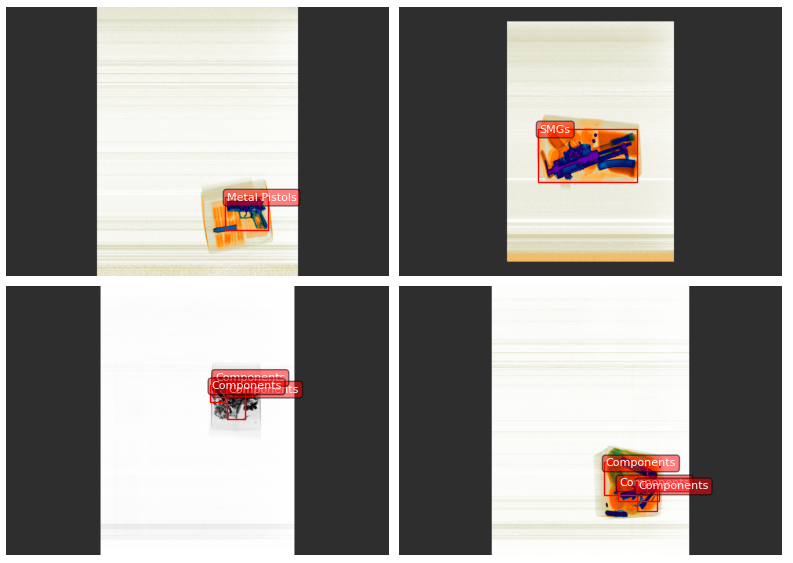}
    \caption{A sample from the CFray dataset.}
    \label{fig:cfray-sample}
\end{figure}

CFray is a custom dataset that contains X-ray scan images of parcels containing firearms and firearm components. The images have been captured using an X-ray scanning machine and annotated using the VoTT \footnote{\href{https://github.com/microsoft/VoTT}{https://github.com/microsoft/VoTT}} annotation tool. Parcels have been filled with both illicit and non-illicit objects to emulate, as closely as possible, real-life illicit parcels. The dataset contains 1368 annotated images, with the supported classes being: ``SMGs", ``Metal Pistols", ``Components", ``Revolvers" and ``Plastic Pistols". Fig. \ref{fig:cfray-sample} depicts indicative reference images.

\subsection{Evaluation Metrics}
This paper presents evaluation results regarding both the achieved object detection accuracy and the achieved inference speed.

\subsubsection{\textbf{Detection Accuracy Metrics}}
Object detectors typically output the bounding boxes/RoIs of the detected objects, the predicted classes, and the corresponding confidence value for each predicted class. Each input image may contain multiple objects (e.g., gun components, gun ammunition, and a gun in one parcel), therefore each object should be separately located and then recognized.

The Intersection over Union (IoU) ratio is calculated by comparing the predicted bounding box with the ground-truth RoI. Ideally, they should be totally overlapping:
\begin{equation}
\text{IoU}=\frac{\text{size of intersection}}{\text{size of union}}.
\end{equation}
However, the calculated IoU output is typically not 1. Therefore, a threshold is used to determine whether the predicted RoI is a sufficiently accurate result. Thresholds of 0.5 and 0.5-0.95 are used in this paper.

The IoU is used as a threshold ratio in the mean Average Precision (mAP) metric, which is one of the evaluation metrics used. To calculate the mAP, true positives (TP), false positives (FP) and false negatives (FN) are also required. Precision measures the percentage of correct predictions, while Recall measures the correct predictions relative to the ground-truth. These metrics are calculated using the following equations:
\begin{equation}
    \text{Precision}=\frac{\text{TP}}{\text{TP}+\text{FP}}.
\end{equation}
\begin{equation}
    \text{Recall}=\frac{\text{TP}}{\text{TP}+\text{FN}}.
\end{equation}
The Precision-Recall (PR) curve illustrates the balance between precision and recall for a given class. Average Precision (AP) is calculated as the area under the PR curve, ranging from 0 to 1. The formula for AP is as follows:

\begin{equation}
    \text{AP}=\int_0^1{p(r)dr}.
\end{equation}
Finally, the mAP is calculated as the mean value of AP for each class over all the N classes:

\begin{equation}
    \text{mAP}=\frac{1}{N}\sum_k^N{\text{AP}_k}.
\end{equation}

\subsubsection{\textbf{Speed Metric}}

Inference time in milliseconds (ms) is a metric that calculates the time required for an object detection method to output the bounding boxes, predicted classes, and confidence value of one input image. It is typically calculated as an average over multiple images.

\subsection{Experimental Setup}
The methods presented in Section \ref{obj-det-methods} have been adopted and compared in the chosen datasets. Various combinations of detectors and backbones have been exploited, focusing as much as possible on comparing Transformer/hybrid components. Thus, both DINO and even the YOLOv8 detection head itself have been combined with the NextViT bacbbone (although YOLO itself is a CNN).

Regarding the connection of RT-DETR and YOLOv8 detectors with the NextViT backbone, i.e., combinations which have been engineered from scratch in this paper, a validation process has been utilized to identify the optimal residual connections. Experiments have shown that the best option for YOLOv8 is to be connected to backbone layers 7 and 17, while the respective result for RT-DETR are layers 9 and 19. However, the final performance differences are very small compared to different connectivity configurations.

All experiments have been conducted on a PC with Ubuntu Linux 22.04, utilizing a 13th Gen Intel(R) Core(TM) i9-13900K CPU and two NVIDIA GeForce RTX 4070 Ti GPUs. Development has taken place in Python, PyTorch and Ultralytics. The training hyperparameters used with the NextViT backbone are the following ones: AdamW \cite{loshchilovDecoupledWeightDecay2019} optimizer with 0.01 learning rate, weight decay of 0.0005 and finally trained for 100 epochs using early-stopping. For SWIN and DINO, the hyperparameter values given in \cite{zhangDINODETRImproved2022} have been adopted.

\subsection{Evaluation Results}
Table \ref{tab:results} shows the results of pretraining the chosen methods on SIXray10-D and finetuning on CFray. All methods are comparable in terms of mAP. However, both methods using the NextViT backbone require about a tenth of the inference time of the SWIN-b and DINO methods, while generating results of similar accuracy. In terms of mAP@50 performance, SWIN+DINO is the most powerful method. In X-rays, it is important to have good mAP scores at all thresholds, hence the more stringent mAP@50-95 threshold evaluation. For this threshold, YOLOv8 \cite{jocherUltralyticsYOLOv82023} combined with NextViT \cite{liNextViTNextGeneration2022} is slightly ahead of SWIN+DINO.

Thus, if real-time performance is needed, the combination of NextViT+YOLOv8 is without question the best choice. On the other hand, if the most important priority is to avoid false negatives (as in many security inspection use-cases) and inference speed is not equally significant, then the SWIN+DINO combination should be considered.

\renewcommand{\arraystretch}{1.2} 
{\rowcolors{3}{gray!20}{}
\begin{table}[t]
    \centering
    \caption{Experimental evaluation results for the selected method combinations (pretraining on SIXray10-D, downstream finetuning on CFRay).}
    \begin{tabular}{|c|c|c|c|c|}
        \hline
        \multicolumn{2}{|c|}{Configuration}
        &
        \multicolumn{3}{|c|}{Metrics} \\
        \hline
        Backbone & Detector & mAP@50 & mAP@50-95 & Inf. Time \\ [0.5ex] 
        \hline
        \fontsize{9}{8}\selectfont SWIN-b  & DINO      & \textbf{0.993}  & 0.816  & 175.4  \\ [0.5ex]
        \fontsize{9}{8}\selectfont NextViT-s & RT-DETR-l   & 0.943  & 0.765  &  16 \\ [0.5ex]
        \fontsize{9}{8}\selectfont NextViT-s & YOLOv8-l    & 0.986  & \textbf{0.82}  &  \textbf{11.6} \\ [0.5ex]
        \hline
    \end{tabular}
    \label{tab:results}
\end{table}}

\renewcommand{\arraystretch}{1.2} 
{\rowcolors{3}{gray!20}{}
\begin{table}[t]
    \centering
    \caption{Experimental results of methods trained only on CFray.}
    \begin{tabular}{|c|c|c|c|c|}
        \hline
        \multicolumn{2}{|c|}{Configuration}
        &
        \multicolumn{3}{|c|}{Metrics} \\
        \hline
        \fontsize{9}{8}\selectfont Backbone & Detector & mAP@50 & mAP@50-95 & Inf. Time \\ [0.5ex] 
        \hline
        \fontsize{9}{8}\selectfont SWIN-b  & DINO      & 0.942  & \textbf{0.837}  & 175.4  \\ [0.5ex]
        \fontsize{9}{8}\selectfont NextViT-s & RT-DETR-l   & 0.94  & 0.748  &  16 \\ [0.5ex]
        NextViT-s & YOLOv8-l    & \textbf{0.984}  & 0.76  & \textbf{11.7}  \\ [0.5ex]
        \hline
    \end{tabular}
    \label{tab:results-ceasefire}
\end{table}}

\renewcommand{\arraystretch}{1.2} 
{\rowcolors{3}{gray!20}{}
\begin{table}[h!]
    \centering
    \caption{Experimental evaluation results for the selected method combinations on SIXray10-D.}
    \begin{tabular}{|c|c|c|c|c|}
        \hline
        \multicolumn{2}{|c|}{Configuration}
        &
        \multicolumn{3}{|c|}{Metrics} \\
        \hline
        \fontsize{9}{8}\selectfont Backbone & Detector & mAP@50 & mAP@50-95 & Inf. Time \\ [0.5ex] 
        \hline
        \fontsize{9}{8}\selectfont SWIN-b    & DINO        & 0.902  & 0.765  & 161.3  \\ [0.5ex]
        \fontsize{9}{8}\selectfont NextViT-s & RT-DETR-l   & 0.889  & 0.762  &  16 \\ [0.5ex]
        NextViT-s & YOLOv8-l    & \textbf{0.906}  & \textbf{0.793}  &  \textbf{12.8} \\ [0.5ex]
        \hline
    \end{tabular}
    \label{tab:results-sixray}
\end{table}}

\subsection{Ablation Studies}
Besides the pretraining+downstream finetuning arrangement, additional experiments have been conducted with single-pass training only on SIXray10-D or on CFray. The respective results are shown in Tables \ref{tab:results-sixray} and \ref{tab:results-ceasefire}, respectively.

Tables \ref{tab:results} and \ref{tab:results-ceasefire} demonstrate that pretraining on the SIXray10-D dataset and then finetuning on CFRay can significantly and clearly improve the achieved accuracy only for NextViT-s+YOLOv8-l, and in fact only when the stricter IoU thresholds are considered.

On the other hand, Table \ref{tab:results-sixray} displays the performance of the employed methods on the SIXray10-D dataset. The NextViT-s+YOLOv8 method outperforms all others on all metrics, especially on the mAP@50-95 metric. The SWIN-b+DINO combination, despite its higher model complexity (and, therefore, lower inference speed) lags in comparison, accuracy-wise. Given that SIXray10-D is a more reasonably sized dataset compared to the very small CFRay dataset, this implies that an end-to-end Transformer architecture faces issues in the highly specific X-ray domain, compared to hybrid solutions that also employ convolutions, but is surprisingly accurate in the low-data regime as shown by Table \ref{tab:results-ceasefire}.

\section{Conclusions}
\label{conclusion}

Automated illicit object detection in X-ray images taken at critical terminals, such as airports, train stations, subways, and ports, is a crucial task for public safety. This paper evaluates efficient and performant methods for handling the large volume and throughput of parcels, luggage, and passengers. The results demonstrate the remarkable accuracy of the DINO Transformer detector in the low-data regime, the impressive real-time performance of YOLOv8, and the effectiveness of the NextViT backbone. Future research directions may involve combining X-ray-specific auxiliary neural modules with the methods evaluated in this paper to further improve accuracy.

\section{Acknowledgment}

The research leading to these results has received funding from the European Union’s Horizon Europe research and innovation programme under grant agreement No 101073876 (Ceasefire). This publication reflects only the authors views. The European Union is not liable for any use that may be made of the information contained therein.

\bibliography{ObjDetXrayViT}
\bibliographystyle{ieeetr}

\end{document}